\documentclass[letterpaper, 10 pt, journal, twoside]{IEEEtran}
\usepackage{epsfig}
\usepackage{epstopdf}
\usepackage{graphics}
\usepackage{epstopdf}
\usepackage{upgreek}
\usepackage{amssymb}
\usepackage{amsfonts}
\usepackage{amsmath}
\usepackage{refstyle}
\usepackage{xcolor}
\usepackage{subfigure}
\usepackage{todonotes}
\usepackage{mathtools}
\usepackage{algorithm}
\usepackage{algpseudocode}
\usepackage{cite}

\newcommand*{\Revise}{\textcolor{black}}

\ifCLASSINFOpdf
\else
\fi
\begin{document}
%
% paper title
% Titles are generally capitalized except for words such as a, an, and, as,
% at, but, by, for, in, nor, of, on, or, the, to and up, which are usually
% not capitalized unless they are the first or last word of the title.
% Linebreaks \\ can be used within to get better formatting as desired.
% Do not put math or special symbols in the title.

\title{Towards Robotic Eye Surgery: Marker-free, Online Hand-eye Calibration using Optical Coherence Tomography Images}
\author{Mingchuan Zhou*, Mahdi Hamad*, Jakob Weiss, Abouzar Eslami, \\Kai Huang, Mathias Maier, Chris P. Lohmann, Nassir Navab, Alois Knoll, M. Ali Nasseri%
% <-this % stops a space
\thanks{Manuscript received: February, 24, 2018; Revised May, 11, 2018; Accepted July, 9, 2018. This paper was recommended for publication by Editor Francois Chaumette upon evaluation of the Associate Editor and Reviewers' comments.
} %Use only for final RAL version This work was supported by (organizations/grants which supported the work.) (Corresponding author: Mingchuan Zhou.)
\thanks{*The first two authors contribute equally to this paper.}
\thanks{Mingchuan Zhou, Mahdi Hamad, and Alois Knoll are with Chair for Robotics and Embedded Systems, Technische Universit\"{a}t M\"{u}nchen, Germany. \{{\tt\small zhoum, mahdi.hamad, knoll\} @tum.de}
		}%		
\thanks{Jakob Weiss and Nassir Navab are with Chair for Computer Aided Medical Procedures and Augmented Reality, Technische Universit\"{a}t M\"{u}nchen, Germany. \{{\tt\small jakob.weiss, navab\} @tum.de}
        }
\thanks{Abouzar Eslami is with Carl Zeiss Meditec, Germany. {\tt\small abouzar.eslami@zeiss.com}
        }%
\thanks{Kai Huang is with School of Data and Computer Science, Sun Yat-Sen University, China. {\tt\small huangk36@mail.sysu.edu.cn}
        }%
\thanks{Mathias Maier, Chris P. Lohmann, and M. Ali Nasseri are with Augenklinik und Poliklinik, Klinikum rechts der Isar der Technische Universit\"{a}t M\"{u}nchen, Germany.
\{{\tt\small mathias.maier, chris.lohmann,ali.nasseri\} @mri.tum.de}
        }%
\thanks{Digital Object Identifier (DOI): 10.1109/LRA.2018.2858744}
}

% note the % following the last \IEEEmembership and also \thanks -
% these prevent an unwanted space from occurring between the last author name
% and the end of the author line. i.e., if you had this:
%
% \author{....lastname \thanks{...} \thanks{...} }
%                     ^------------^------------^----Do not want these spaces!
%
% a space would be appended to the last name and could cause every name on that
% line to be shifted left slightly. This is one of those "LaTeX things". For
% instance, "\textbf{A} \textbf{B}" will typeset as "A B" not "AB". To get
% "AB" then you have to do: "\textbf{A}\textbf{B}"
% \thanks is no different in this regard, so shield the last } of each \thanks
% that ends a line with a % and do not let a space in before the next \thanks.
% Spaces after \IEEEmembership other than the last one are OK (and needed) as
% you are supposed to have spaces between the names. For what it is worth,
% this is a minor point as most people would not even notice if the said evil
% space somehow managed to creep in.

% The paper headers
%\markboth{Journal of \LaTeX\ Class Files,~Vol.~14, No.~8, August~2015}%
%{Shell \MakeLowercase{\textit{et al.}}: Bare Demo of IEEEtran.cls for IEEE Journals}
\markboth{IEEE Robotics and Automation Letters. Preprint Version. Accepted July, 2018}
{ZHOU \MakeLowercase{\textit{et al.}}: Towards Robotic Eye Surgery: Marker-free, Online Hand-eye Calibration}

% The only time the second header will appear is for the odd numbered pages
% after the title page when using the twoside option.
%
% *** Note that you probably will NOT want to include the author's ***
% *** name in the headers of peer review papers.                   ***
% You can use \ifCLASSOPTIONpeerreview for conditional compilation if
% you desire.

% If you want to put a publisher's ID mark on the page you can do it like
% this:
%\IEEEpubid{0000--0000/00\$00.00~\copyright~2015 IEEE}
% Remember, if you use this you must call \IEEEpubidadjcol in the second
% column for its text to clear the IEEEpubid mark.

% use for special paper notices
%\IEEEspecialpapernotice{(Invited Paper)}

% make the title area
\maketitle

% As a general rule, do not put math, special symbols or citations
% in the abstract or keywords.
\begin{abstract}
Ophthalmic microsurgery is known to be a challenging operation, \Revise{which requires very precise and dexterous manipulation}. Image guided robot-assisted surgery (RAS) is a promising solution that brings significant improvements in outcomes and reduces the physical limitations of human surgeons. However, this technology must be further developed before it can be routinely used in clinics. One of the problems is the lack of proper calibration between the robotic manipulator and appropriate imaging device. In this work, we developed a flexible framework for hand-eye calibration of an ophthalmic robot with a microscope-integrated Optical Coherence Tomography (MI-OCT) without any markers. The proposed method consists of three main steps: a) we estimate the OCT calibration parameters; b) with micro-scale displacements controlled by the robot, we detect and segment the needle tip in 3D-OCT volume; c) we find the transformation between the coordinate system of the OCT camera and the coordinate system of the robot. We verified the capability of our framework in ex-vivo pig eye experiments and compared the results with a reference method (marker-based). In all experiments, our method showed a small difference from the marker based method, with a mean calibration error of 9.2 $\upmu$m and 7.0 $\upmu$m, respectively. Additionally, the noise test shows the \Revise{robustness} of the proposed method.
\end{abstract}

% Note that keywords are not normally used for peerreview papers.
% \begin{IEEEkeywords}
% IEEE, IEEEtran, journal, \LaTeX, paper, template.
% \end{IEEEkeywords}

% For peer review papers, you can put extra information on the cover
% page as needed:
% \ifCLASSOPTIONpeerreview
% \begin{center} \bfseries EDICS Category: 3-BBND \end{center}
% \fi
%
% For peerreview papers, this IEEEtran command inserts a page break and
% creates the second title. It will be ignored for other modes.
\IEEEpeerreviewmaketitle

\section{Introduction}
\label{intro}
\begin{figure}[htbp]
  \subfigure[Ophthalmic operation]{
   \centering \includegraphics[width=0.45\columnwidth,height=0.12\textheight]{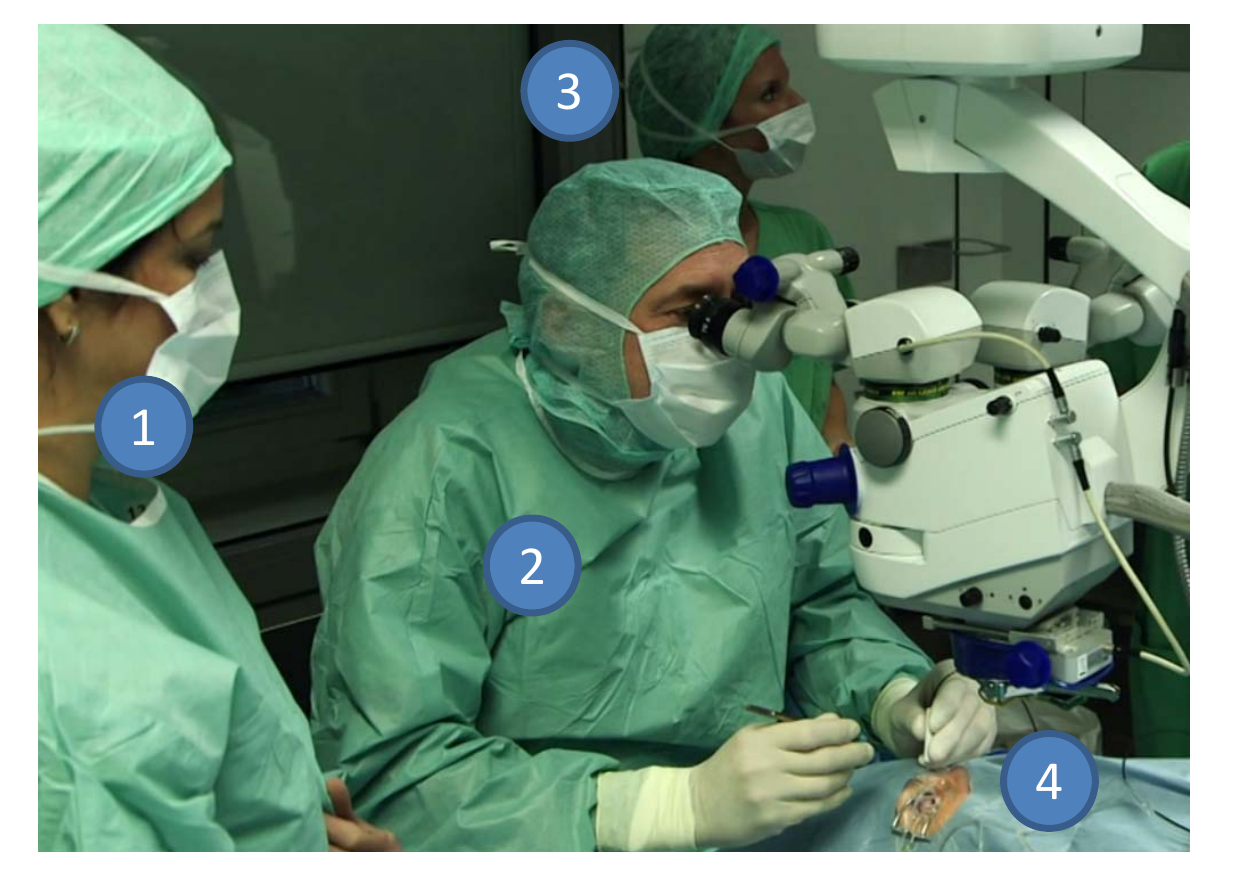}
    \label{fig:Introduction_a}
  }
  \subfigure[Microscope view of focal region]{
    \includegraphics[width=0.45\columnwidth,height=0.12\textheight]{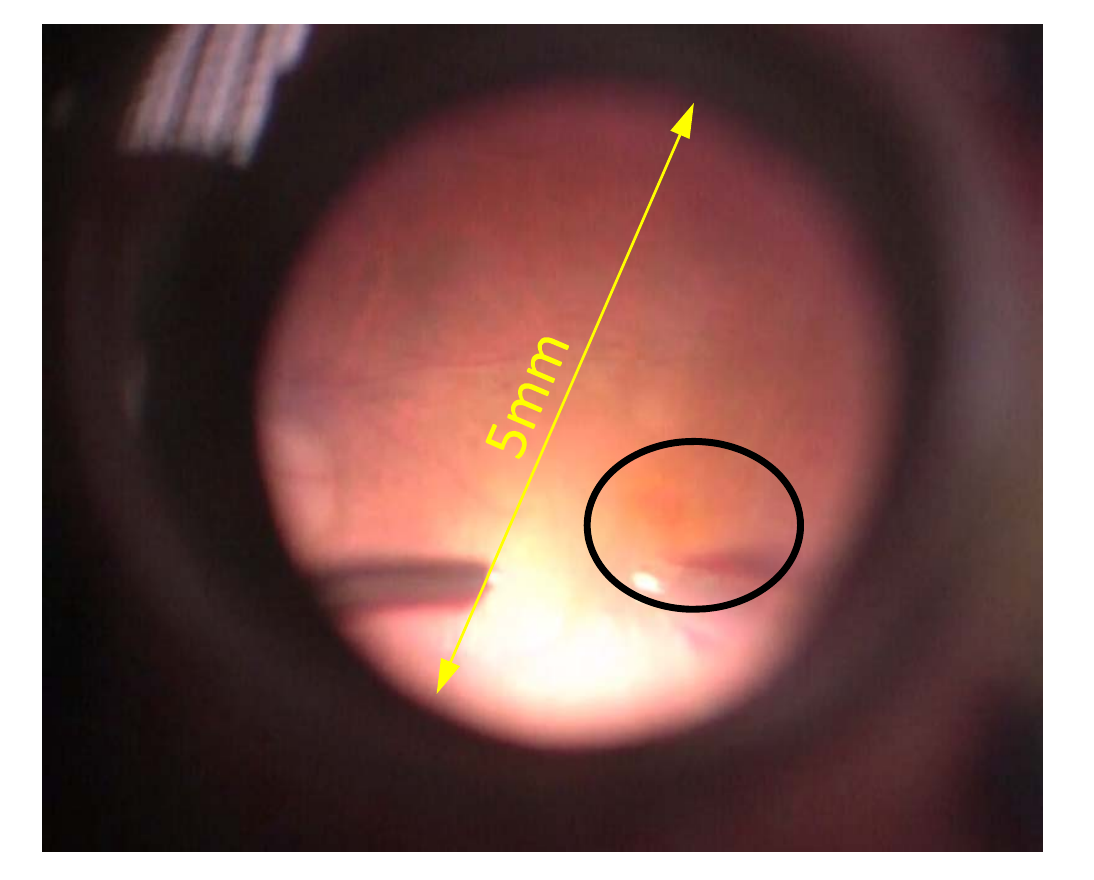}
    \label{fig:Introduction_b}
  }
  \caption{(a): A conventional setting in an ophthalmic operation room: a surgeon (2); two assistants (1) and (3); the patient (4). (b): The focal region in the microscopic view. The operation area is marked by an ellipse.}
  \label{fig:introsurg}
\end{figure}

\IEEEPARstart{O}{phthalmic} surgery is a typical microsurgery with a delicate and complex workflow, which needs critical surgical skills and considerations (Fig.~\ref{fig:introsurg}).  Due to this fact, the ophthalmologists are required not only to have enough training and clinical experiences but also need to be in a good physical condition to cope with hand tremors and \Revise{a non-stop surgery session}. However, for surgeons, \Revise{clinical experience normally comes with
age, but aging impairs physical skills}. Recently, robot-assisted surgery (RAS) has been known as a solution for reducing the work intensity, increasing the surgical outcomes and prolonging the service time of experienced surgeons. To address the challenges of ophthalmic surgery, many researchers in the past decade introduced high precision robotic setups with different scales and design mechanisms~\cite{nakano2009parallel,ang2001design,wei2007design,taylor1999steady,ullrich2013mobility,meenink2010master,rahimy2013robot,song2012active,nasseri2013introduction,gijbels2014experimental}. Over the years, these robots \Revise{are getting closer to clinical trials}. In September 2016, surgeons at Oxford's John Radcliffe Hospital, performed the \Revise{world's} first robotic Internal Limiting Membrane (ILM) peeling, \Revise{in which the surgeon removes one membrane from a specific area of the retina. The eye surgical robot named} Robotic Retinal Dissection Device (R2D2) \Revise{with 10 $\upmu$m accuracy was used in the clinical trials}~\cite{meenink2013robot}\cite{gonenc20173}, which proved the feasibility of RAS application in one of the most difficult ophthalmic operations.

\Revise{Furthermore, advances in ophthalmologic imaging modalities and sensing devices are
always being introduced to the surgery, which, when integrated with robotic systems,
will give surgeons more confidence to make reliable intra-operative decisions}. %Despite the successful results of robotic applications in clinical trials, in order to fully integrate these machines into clinical routines, more dedicated sensing information is required. More sensory feedbacks give surgeons confidence to make reliable intra-operative decisions. One of the most important sources is visual feedback.%
In addition to conventional en-face images from the ophthalmic microscopes, Optical Coherence Tomography (OCT) is currently known to be a proper imaging modality for visualizing the micro structural anatomy of the eye~\cite{huang1991optical}, as it offers suitable resolution and non-invasive radiation. Compared to the stereo-microscope vision system proposed by Probst et al.~\cite{probst2018automatic}, the drawbacks of OCT are expensive hardware, acquisition and processing time, while the advantages are (a) the imaging quality will not be influenced by the illumination, which in some cases (\Revise{e.g. vitreoretinal surgery, an operation on the posterior segment of the eye}) is not easy to control, and (b) the OCT can provide information inside the tissue, which allows surgeons to analyze the anatomical structures. The latest OCT machines have been developed to even give real-time information of the interior and posterior segment of the eye as well as the interactions between surgical instruments and intraocular tissues~\cite{ehlers2014intraoperative}.

Some research groups integrated the OCT probe into surgical instruments for navigation in head surgery~\cite{diaz2013towards} and ophthalmic surgery~\cite{cheon2015active}. However, to localize the global location of a probe, this may need preoperative Computer Tomography (CT) or Magnetic Resonance Imaging (MRI). These extra preoperative images and modified surgical instruments may increase the complexity of the application.

Instead of using OCT integrated instruments, we propose using the microscope-integrated OCT (MI-OCT) for RAS navigation. In MI-OCT, \Revise{the OCT engine is integrated inside the microscope head and} shares the same optical path with the microscope in order to capture volumetric images, thus the OCT probe avoids integration into surgical instrument. During the planned MI-OCT guided RAS, the surgeon will move the microscope head and adjust its view to locate the focal region (Fig.~\ref{fig:Introduction_b}). Usually, the microscope imaging area is enough to cover the focal area, thus the microscope head could be fixed. \Revise{Following this fixation,} the surgeon will manipulate the robot and move the needle close to the focal area. Therefore, hand-eye calibration is essential for such applications e.g. keeping a safe distance between needle tip and retina or performing path planning for targeted drug delivery.

\begin{figure*}[!htbp]
%\vspace{0.1cm}
\label{sec:method}
\centering
  \includegraphics[width=0.67\textwidth]{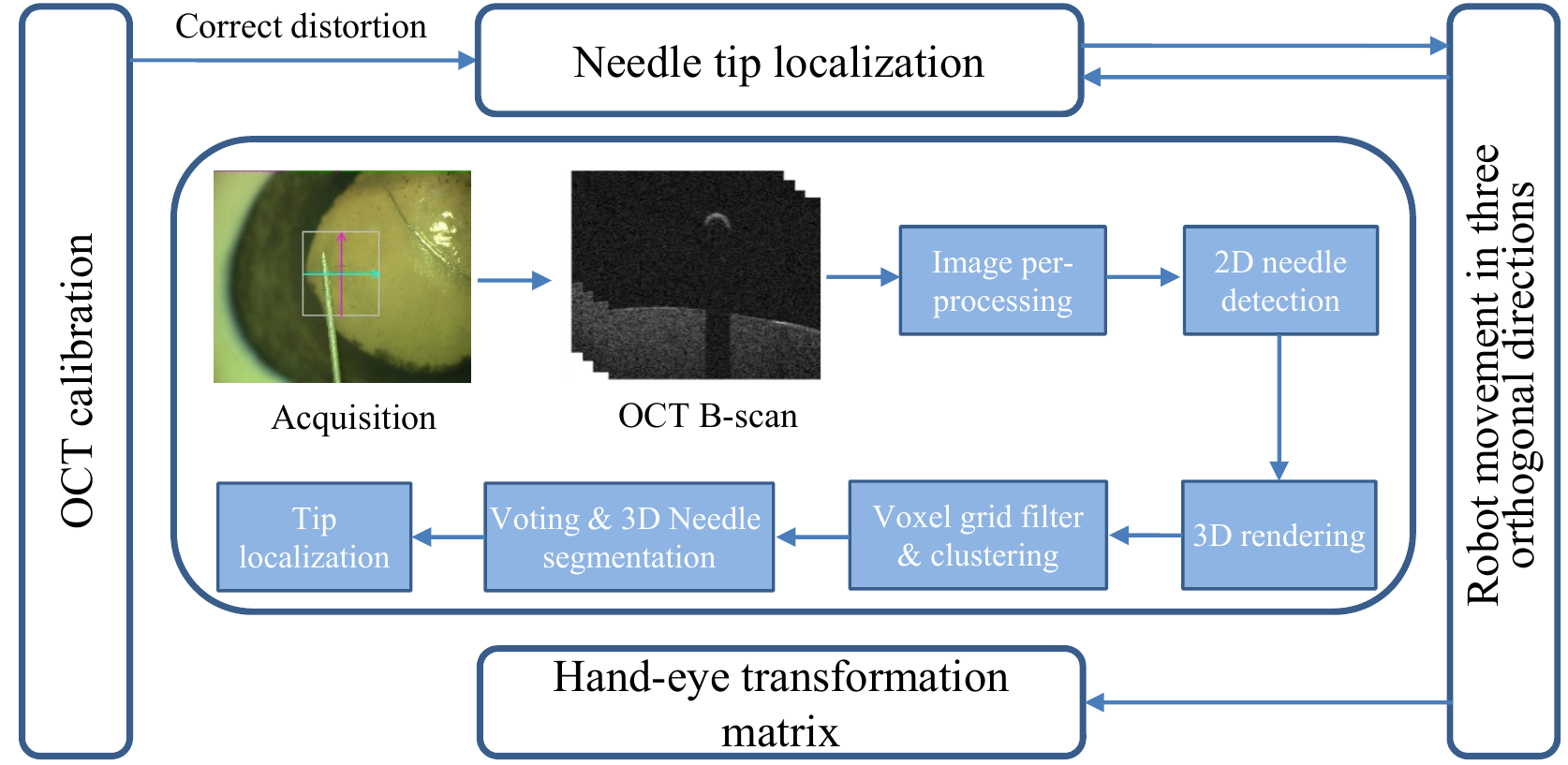}
% figure caption is below the figure
\caption{The framework of the proposed method.}
\label{fig:framework}       % Give a unique label
\end{figure*}

In this paper, we address the problem of hand-eye calibration for a surgical robot with the MI-OCT camera. For this purpose, we use a surgical robot developed by~\cite{nasseri2017targeted}. In this setup, \Revise{head} and eye fixations are used to reduce the risk of movement caused by the patient. The imaging device and the robot should also be rigidly connected to the operation bed following the configuration of a da Vinci surgical system. These setups are necessary for the RAS surgery and will usually require only a one-time calibration. To the best of our knowledge, this is the first paper that proposes such a calibration framework. The main contributions of this paper are summarized as follows:
\begin{itemize}
\item  We propose a flexible framework for hand-eye calibration of an ophthalmic robot and OCT camera without using additional markers;
\item  A robust method for needle tip segmentation and localization in OCT volume is proposed;
\item  Clinical grade accuracy performance (9.2 $\upmu$m) of the calibration method is demonstrated by experimental evaluations on ex-vivo pig eyes.
\end{itemize}
The remainder of the paper is organized as follows. In the next section, we briefly present related work.
The proposed hand-eye calibration framework is described in Section~\ref{sec:method}.
In Section~\ref{sec:experiment}, the performance of the proposed method is evaluated. Experimental results are discussed in Section~\ref{sec:results}.
 Finally, Section~\ref{sec:conculsion} concludes this paper and presents the future work.

\section{Related Work}
\label{sec:relatedwork}
\Revise{Hand-eye calibration is an active research topic in robotics and has been investigated by many researchers. In particular, the seminal work of Tsai et al. since 1980s~\cite{tsai1989new}. There exist several off-the-shelf calibration solutions that implement Tsai's method (e.g. ViSP library~\cite{marchand2005visp}, which is designed for visual servoing applications). These existing off-the-shelf calibration solutions can be easily integrated into standard camera vision systems and allow fast prototyping. However, they require separate measurements and extra equipment (calibration grids, markers,...). In some medical scenarios, it is not practical to place the calibration objects or markers inside the tissue or bodily organ. In addition, medical imaging modalities (e.g OCT and ultrasound) have different imaging principles, which leads to a different camera calibration process.}

In the surgical environment, hand-eye calibration is normally transferred into a Procrustean problem by the~\cite{shiu1989calibration} and has been modified to meet with different surgical requirements in various 3D medical imaging scenarios, e.g. two-camera stereoscopic vision and ultrasound. To determine the unknown fixed transformation between the robot and imaging coordinates, a ball-tip stylus or chessboard is used as a calibrator for hand-eye calibration~\cite{morgan2017hand}.  However, these markers may influence the conventional surgical workflow as well as cause complication \Revise{with} sterility. Thompson et al.~\cite{thompson2016hand} designed a tracking collar for rigid laparoscopes hand-eye calibration. The method can achieve 0.85 mm error, and the calibration process can be finished within 3 minutes. In another work, 3D markers were used by Bergmeir et al.~\cite{bergmeir2009comparing} to realize \Revise{ultrasound} probe calibration with a robotic arm. In terms of none marker research work, Sarrazin et al.~\cite{sarrazin2015hand} used \Revise{ultrasound} volume registration data to achieve calibration that does not require 3D localizers. Pachtrachai et al.~\cite{pachtrachai2016hand} proposed a calibration method where an instrument with known geometry is used instead of an additional calibrator. The calibration was achieved without compromising sterility. \Revise{Francisco et al.~\cite{vasconcelos2016similarity,vasconcelos2016spatial} used constraints based on the needle orientation in 2D/3D ultrasound image to calibrate the ultrasound probe.}

Taking the above research work and requirements of ophthalmic surgery into account, we can conclude that the following considerations need to be taken into account for a qualified calibration method: (a) external calibration objects are best avoided to reduce surgical complication; (b) small computational overhead can make the method easier for online implementation; (c) the calibration accuracy needs to be kept within clinical tolerance. Unlike the aforementioned surgical applications, the intended operation area for ophthalmic surgery is very small and the precision requirement is much higher. The precision requirement of ophthalmic surgery varies depending on the specific operation. For the ILM peeling process, the surgeon can do it quite well, which means the 182 $\upmu$m accuracy (the hand tremor RMS
amplitude~\cite{riviere2000study}) is enough. For the sub-retinal injection, the average thickness of retina is around 200 $\upmu$m, therefore, 20 $\upmu$m would be an acceptable position accuracy. For the retinal vein cannulation, the ideal position accuracy would be 20 $\upmu$m, since the diameter of branch retinal veins is typically less than 200 $\upmu$m. Therefore, the typical position accuracy for robotic eye surgery is considered to be roughly around 10 $\upmu$m~\cite{ang2001design,rahimy2013robot,gijbels2014experimental} to meet all potential operations.

Due to the properties of microsurgeries, it will be complicated to properly place calibration markers during surgery, specifically for the posterior parts, where the markers could move inside the eye. To overcome the high accuracy requirement challenge, we propose using the robot's precise movement, in three \Revise{orthogonal} directions, and the detected needle tip position, to obtain the transformation matrix between the camera and the robot coordinate system.
%Currently, to the best of our knowledge, there is no standard for the accuracy of ophthalmic surgery.

\section{Method}
The proposed framework is shown in Fig.~\ref{fig:framework}. The OCT calibration parameters are obtained in advance to eliminate the distortion caused by the imaging device. With micro-scale displacements controlled by the robot, we move the needle in a trajectory that contains three \Revise{orthogonal} directions. Following each movement, an OCT volume is captured, and the needle tip is detected and accurately segmented even with the presence of image \Revise{noise}. After finishing all the movements defined by the trajectory, we calculate the transformation between the coordinate system of the OCT camera and the coordinate system of the robot. The details of each step are presented as follows.

\subsection{OCT calibration}
Our setup contains an OPMI LUMERA 700 with integrated RESCAN 700 OCT (Carl Zeiss Meditec AG., Germany). \Revise{The OCT engine is spectral domain OCT, which has improved scanning speed compared to time domain OCT. The OCT engine has a wavelength of 840 nm and a scanning speed of 54000 A-scans per second.} Due to the fact that the intended surgical area for ophthalmic operation is usually very small, we set the internal mirror galvanometer, \Revise{which can deflect a light beam by an electric current}, in the OCT to obtain the scan area 3.01 mm$\times$3.10 mm$\times$2.60 mm with the highest resolution of 128 B-scans, each with 512 A-scans. \Revise{B-scan refers to a two dimensional, cross-section scan and A-scan is one dimensional scan.}
% Thus we use the method proposed by Van der Jeught et. al to estimate the distortion of OCT.

\begin{figure}[!htbp]
  \includegraphics[width=0.95\columnwidth]{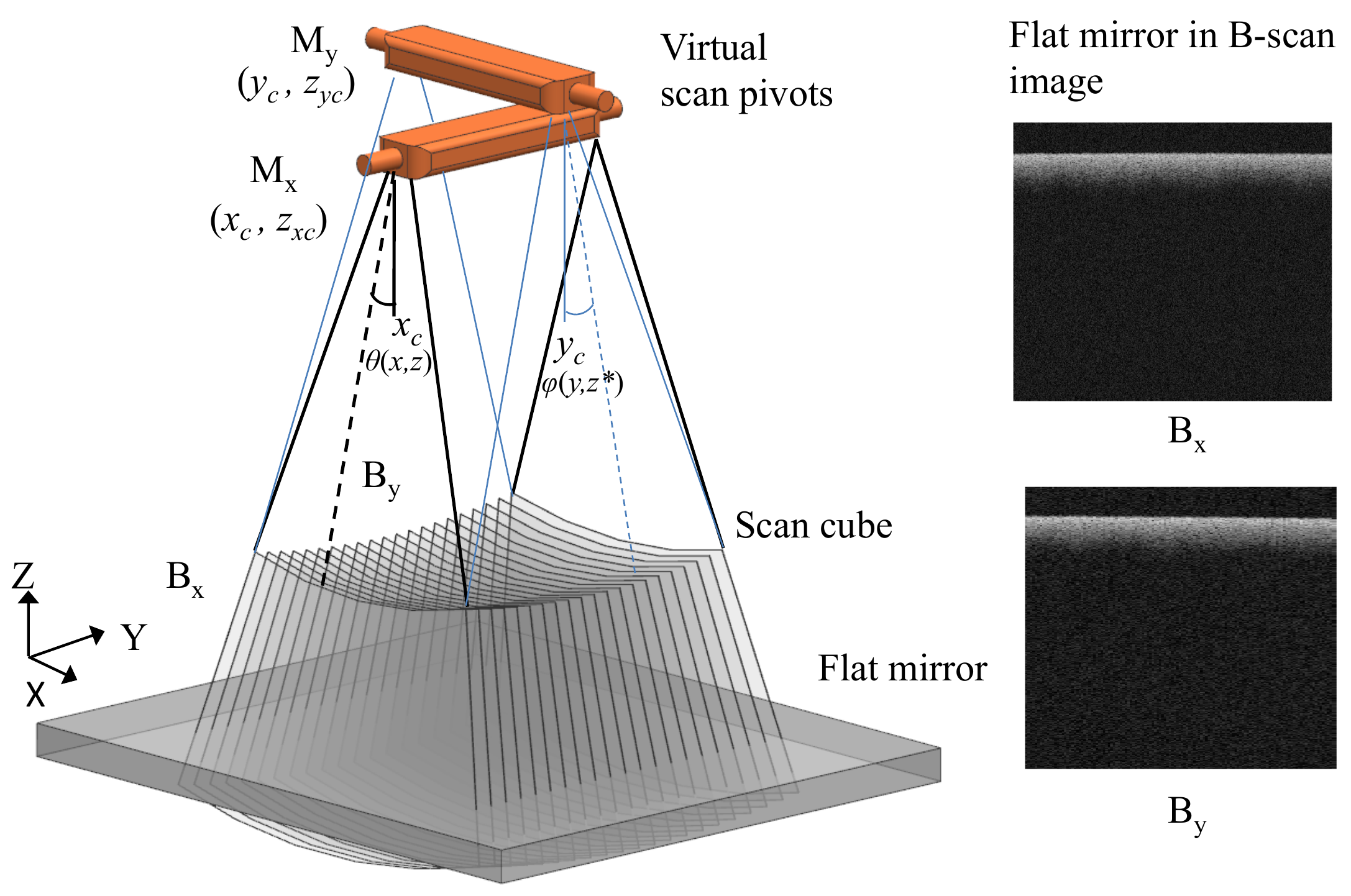}
% figure caption is below the figure
\caption{The distortion of OCT scan and the cross-sectional B-scan images.}
\label{fig:OCTscan}       % Give a unique label
\end{figure}

The simplified principle of an OCT scanner is shown in Fig.~\ref{fig:OCTscan}. \Revise{The cross-sectional profiles of the flat surface is curved, because the two scan mirrors $M_x$ and $M_y$ have different scan radius which arch the beaming path}. During the scanning in the X- and Y- direction,  $M_y$ and $M_x$ are fixed, respectively. Therefore, there are geometric distortion effects imposed onto the OCT volume by two \Revise{mirror galvanometers}, which are considered to be decoupled from each other. We refer to the method which is proposed by Van der Jeught et al.~\cite{van2013real} to correct the OCT distortion. The glass slider (26 mm$\times$76 mm$\times$1 mm) with flatness error of $\pm$3 $\upmu$m is used as the flat surface (Brand GmbH, Germany). The topmost surface of the mirror is recognized with the threshold filter in each B-scan image. The radius of the topmost surface is fitted by the analytical equation of a circle, thereby the virtual pivot center $(x_c, z_{xc})$ of the \Revise{mirror galvanometer} is calculated by averaging all fitted circle centers for the B-scans in X-direction ($B_x$). We repeat the same process to obtain the virtual pivot center $(y_c, z_{yc})$ with all the B-scans in Y-direction ($B_y$). The distortion correction equation is \Revise{reported below},
\begin{gather}
\label{equa:basic1}
\begin{cases}
x' = R_x(x, z)sin(\theta(x,z))+x_c\\
y' = R_y(y,z^*)sin(\varphi(y,z^*))+y_c\\
z' = R_y(y,z^*)cos(\varphi(y,z^*))+z_{yc}\\
\end{cases}
\end{gather} where the point $(x', y', z')$ is the corrected result of the point $(x, y, z)$ in the original volume, $R_x(x, y)$ denotes the distance between virtual pivot center $(x_c, z_{xc})$ and $(x, z)$, $sin(\theta(x,z))$ denotes the angle of polar coordinate system with the virtual pivot center of $M_x$, $z^*$ which equals $R_x(x ,z)cos(\theta(x,z))+z_{xc}$ is the corrected $z$ value for the distortion causing from $M_x$ scanning mirror, $R_y(y,z^*)$ denotes the distance between virtual pivot center $(y_c, z_{yc})$ and $(y,z^*)$, $\varphi(y,z^*)$ denotes the angle of the polar coordinate system with the virtual pivot center $M_y$.

%With known the correction parameters in Tab.~\ref{tab:OCTparameters}, we could correct the any point.
%Meanwhile, the maximum of the distortion $d_{max}$ can be estimated based on this calibration model.

\subsection{Needle tip localization}

\begin{figure}[!htb]
\vspace{0.1cm}
  \subfigure[Orginal B-scan image]{\includegraphics[width=0.44\columnwidth,height=0.15\textheight]{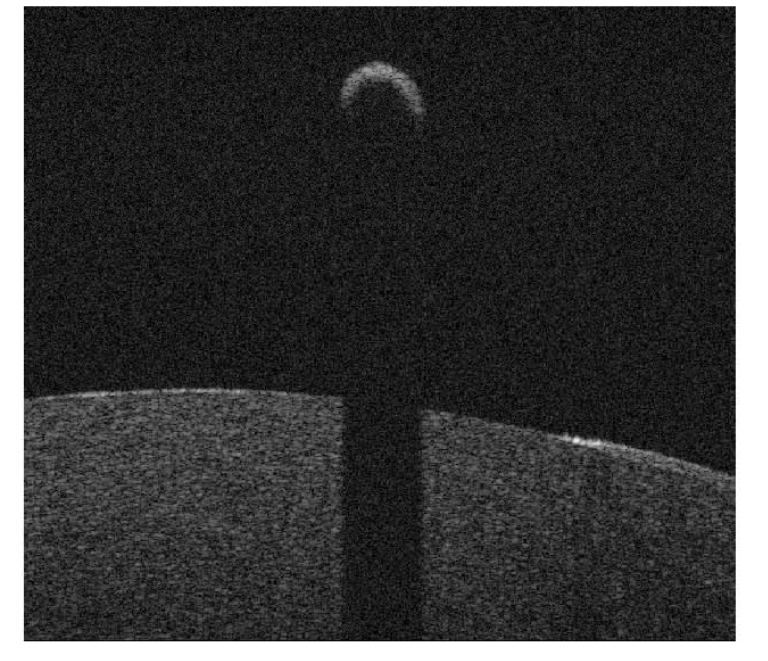}
    \label{fig:ellipsefit_a}
  }
  \subfigure[Needle detection]{
    \includegraphics[width=0.41\columnwidth,height=0.15\textheight]{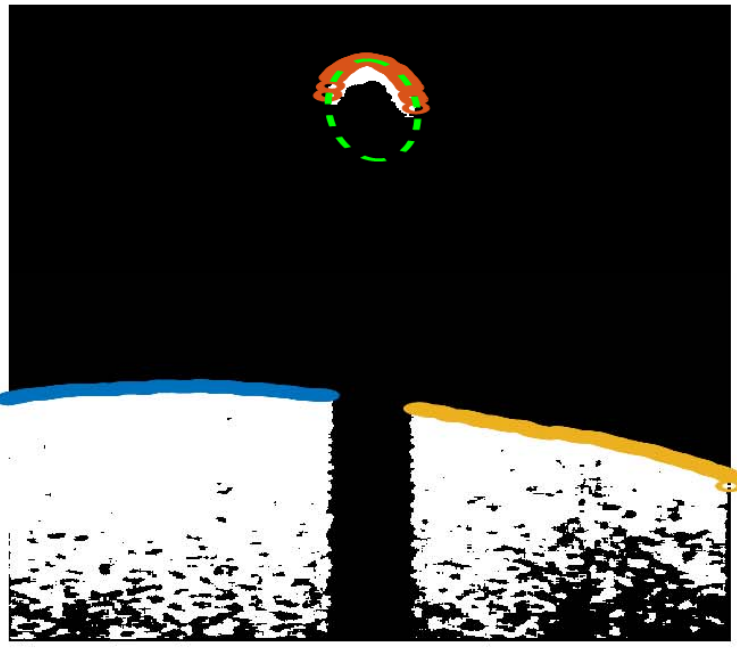}
    \label{fig:ellipsefit_b}
  }
  \caption{The original B-scan image (a) and needle detection using ellipse fitting for topmost of contours (b). The ellipse with green dash line is the detected ellipse for needle body. The contours are marked with \Revise{different colors for different connection areas}.}
  \label{fig:ellipsefit}
\end{figure}

Instead of using 3D markers for hand-eye calibration, in our approach, we directly utilize a robust and precise method to localize the needle tip in OCT images. The \Revise{actual} tip is invisible due to the light diffraction and its optical property, therefore, this detected needle tip is the smallest tip part in the OCT images. The original B-scan gray image (Fig.~\ref{fig:ellipsefit_a}) is transformed into a binary image by the thresholding method. The threshold value is adaptively defined, based on the statistical measurements of each B-scan~\cite{roodaki2015introducing}. We eliminate the noise inside the binary image, by applying a median filter followed by a Gaussian filter. \Revise{The sizes} of the median and Gaussian filters are much less than the visible needle tip. Therefore, the smallest visible tip part will not be missed as a result of the de-noising procedure.

  A voting mechanism is used to specify whether a point in the 2D image belongs to the needle body. Since the needle body part in the B-scan image is a half ellipse, which can be considered as a strong feature, the ellipse fitting is applied to the topmost contours of each B-scan. The topmost contours are obtained by scanning each column of pixels going from top to bottom (Z-direction), starting from the leftmost column and moving to the right (X-direction). The connected contour pixels ( blue, yellow, and red in Fig.~\ref{fig:ellipsefit_b}) are considered as one group and fitted to an ellipse equation. In order to filter any ellipse detected other than the needle body (e.g. blue$\&$yellow eye tissue), we constrain the ellipse's minor axis to a value lower than $m_e$, where $m_e$ is defined based on the diameter of the instrument (i.e. needle). The fitted ellipse is shown as a green color in Fig.~\ref{fig:ellipsefit_b}. Finally, an additional property (\Revise{Boolean} value) is added to each pixel in B-scan. \Revise{The Boolean value is true if the pixel belongs to the fitted ellipse. Otherwise, it is false.}

% To reduce the excessive amounts of data and computation overhead, a voxel grid filter is used. The filter takes all the points in a cloud and downsamples them to a specified resolution. Here, we chose a leaf size of 0.02 mm.

\begin{figure}[!htb]

\centering
  \subfigure[3D OCT volume]{
    \includegraphics[width=0.45\columnwidth]{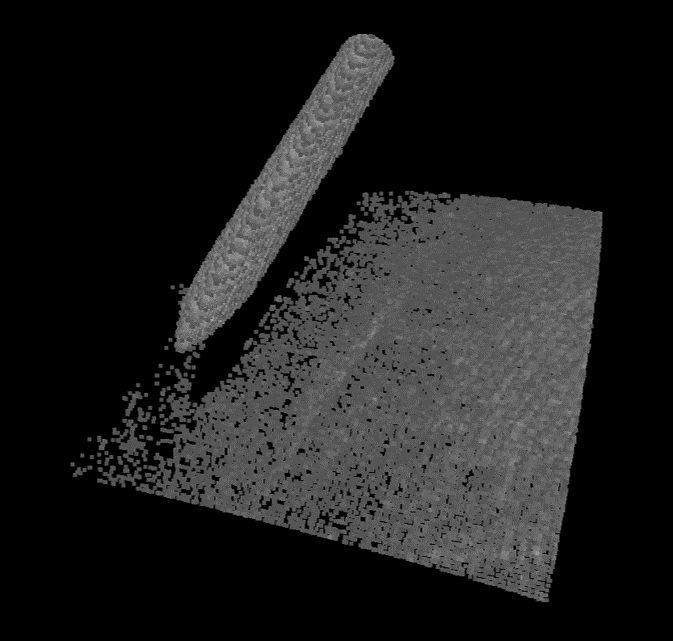}
    \label{fig:Cube_a}
  }
  \subfigure[Segmented point clusters]{
    \includegraphics[width=0.45\columnwidth]{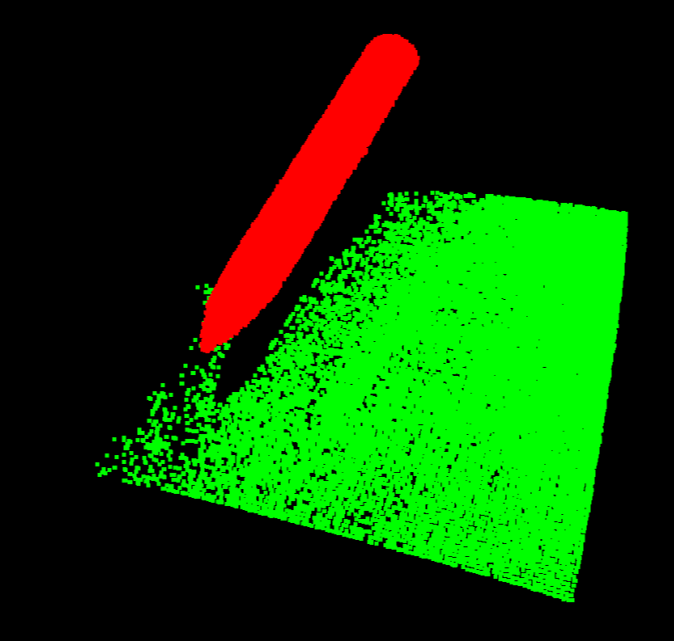}
    \label{fig:Cube_b}
  }
  \subfigure[Voting results]{
    \includegraphics[width=0.45\columnwidth]{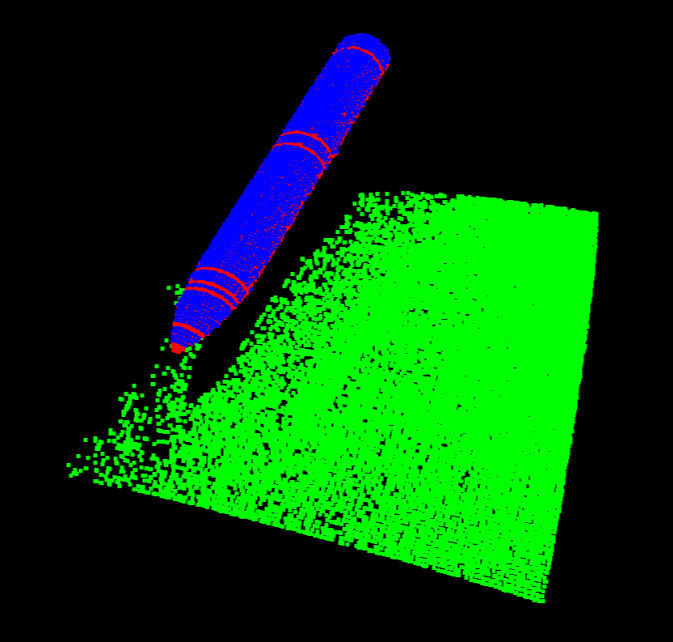}
    \label{fig:Cube_c}
  }
   \subfigure[Needle tip localization]{
    \includegraphics[width=0.45\columnwidth]{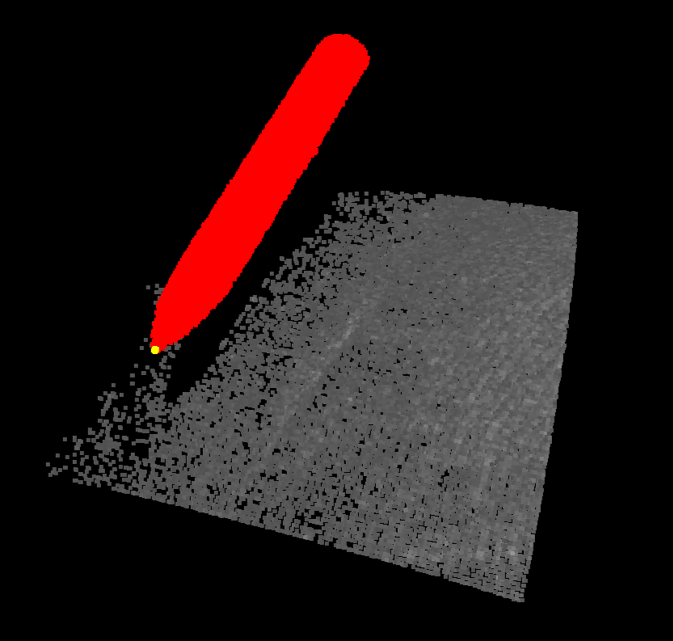}
    \label{fig:Cube_d}
  }
  \caption{(a) The needle with ex-vivo pig eye tissue in 3D OCT volume. (b) The points are clustered and marked with different colors. (c) 2D Detected ellipses from Fig.~\ref{fig:ellipsefit_b} are colored in blue and used for voting. (d) Needle segmentation and needle tip localization (the needle tip is marked by a yellow point).}
  \label{fig:cube}
\end{figure}

In order to render the images in 3D, the OCT volume is represented as a point cloud structure \(\mathcal{P}\) (Fig.~\ref{fig:Cube_a}) with $(x_i, y_i, z_i, b_i)$, where $(x_i,y_i,z_i)$ is the position of point $p_i \in \mathcal{P}$, and $b_i$ is the \Revise{Boolean} value to identify whether the point belongs to the needle. \Revise{To reduce the excessive amounts of data and computation overhead, a voxel grid filter with a leaf size of 0.02 mm is used.} Afterwards, we need to differentiate between objects in the point cloud. For this, we use Euclidean cluster extraction. The clustering step uses a Kd-tree structure for finding the nearest neighbors of every point in the cloud. Two points $p_i$ and $p_j$ belong to two different clusters $c_i$ and $c_j$ if:

\begin{equation} \label{eq:2}
     min \parallel{p_i - p_j}\parallel_2 \geq t
\end{equation}
where $t$ is the maximum imposed distance threshold. The result of these processing steps (Fig.~\ref{fig:Cube_b}) is a set of euclidean point clusters $C = \{(c_1)...(c_n)\}$, where each segmented cluster $c_i$ is at least $t$ far away from the other cluster. At this point, we have two different clusters, but we do not know which one is the needle. To segment the needle (see Algorithm \ref{detect-needle}), the cluster that has the greatest number of points with $b_i$ = 1 (i.e. most voting, blue color in Fig.~\ref{fig:Cube_c}) will be treated as the needle (Fig.~\ref{fig:Cube_d}). In order to localize the needle tip, the B-scan direction can be adjusted manually or automatically to match with the needle insertion direction~\cite{roodaki2015introducing}. \Revise{The alignment will not add extra complexity since it only costs several seconds manually with the scanner control panel of RESCAN 700 OCT.} As a result, the needle tip will always be located in the first B-scan (slice) of the segmented needle. The yellow point (Fig.~\ref{fig:Cube_d}) is the centroid of the needle cluster in this slice, which represents the needle tip. Finally, the needle tip position is calibrated by  Eqn.~\ref{equa:basic1} to correct for the distortion in the OCT volume.

\begin{algorithm}[!htb]
		\caption{Segmentation of needle body in 3D }
		\label{detect-needle}
         \textbf{INPUT:} $C$ - Euclidean point clusters $\{(c_1)...(c_n)\}$\\
        \textbf{OUTPUT:} $c_i$ - Cluster with the greatest number of \Revise{votes}
		\begin{algorithmic}[1]
			\Procedure{SegmentNeedle}{$C$}
            \State $maxVote = 0$, $maxIndex = 0$, $vote = 0$
			\For {each cluster $i$ in $C$}
            \For {each point $j$ in $i$}
            \If {$b_j == 1 $}
            \State $vote = vote +1$
			\EndIf
			\EndFor
            \If {$vote > maxVote$}
            \State $maxVote = vote$
            \State $maxIndex = i$
			\EndIf
            \State $vote = 0$
            \EndFor
			\State Return $C(maxIndex)$
			\EndProcedure
		\end{algorithmic}
\end{algorithm}

% Therefore, for a successful segmentation, the method requires different objects to be at least $t$ far away from each other.
% \begin{figure*}[h]
%   \includegraphics[width=0.8\textwidth]{figures/Cube.pdf}
% % figure caption is below the figure
% \caption{The distortion of OCT scan and the cross-sectional B-scan images.}
% \label{fig:cube}       % Give a unique label
% \end{figure*}
\vspace{-1em}
\subsection{Hand-eye calibration}\label{sec:methodhand}
The hand-eye calibration is usually defined by solving a system of homogeneous matrix equations in the form $AX = ZB$, where $X$ and $Z$ are rigid transformations defined by the configuration used. The calibration usually involves eye-in-hand and eye-to-hand configurations. The camera is said to be eye-in-hand when rigidly mounted on the robot end-effector, and it is said to be eye-to-hand when it observes the robot within its work space. The latter is our setup, and since our needle is attached directly to the end effector of the robot, and given the unique kinematics of the robot~\cite{nasseri2017targeted}, we can find $A' = AX $. We then move the robot with micro-scale displacements and formulate the system as follows:
\begin{align}\label{eq:hand-eyeMatrix}
\begin{bmatrix}A'_i \\ \vdots \\ A'_j\end{bmatrix}  = \begin{bmatrix}R &| & T \end{bmatrix} \times \begin{bmatrix}B_i \\ \vdots \\ B_j\end{bmatrix}
\end{align}

% where $A' = \{A'_i,..., A'_j\}$ is the 3D needle position in the robot coordinate system and $B = \{B_i,...,B_j\}$ is the corresponding needle position in the camera coordinate system. $R$ $\in$ $SO_3$  and $T$ $\in$ $\mathbb{R}^3$ are the rotation matrix and translation vector that transform the needle positions from the OCT camera coordinate system to the robot coordinate system.

where $A' = \{A'_i,..., A'_j\}$ is the 3D needle position in the robot coordinate system and $B = \{B_i,...,B_j\}$ is the corresponding needle position in the camera coordinate system. \Revise{$R$ $\in$ $SO_3$  and $T$ $\in$ $\mathbb{R}^3$ are the rotation matrix and translation vector that transform the needle positions from the OCT camera coordinate system to the robot coordinate system. The transformation matrix is represented as,
\begin{align}\label{eq:matrixT}
\begin{bmatrix}R &| & T \end{bmatrix} = \begin{bmatrix}R_{3 \times 3}  & T_{3 \times 1} \\ 0_{1 \times 3} & 1
\end{bmatrix}
\end{align} }
% The latter is our setup, and is defined by creating a linear system of equation:
%\begin{equation}
%\label{equa:basic2}
%R = XO
%\end{equation}
%where $R$ and $O$  denotes the 3D needle tip position in the robot coordinate system and  OCT coordinate system, respectively. $X$ denotes a 3D \Revise{affine} transformation matrix.%
%We apply two different methods to solve Eqn.~\ref{eq:hand-eyeMatrix}: one is to get transformation matrix based on singular value decomposition~\cite{svd1987} (SVDT); and the other one is using unit quaternions\cite{horn1987} (QT). SVDT calculates the rotation and translation parameters separately while QT uses a unit quaternion instead of rotation.%
\Revise{We apply two different methods to solve Eqn.~\ref{eq:hand-eyeMatrix}: in the first method (SVDT), we calculate the rotation and translation parameters separately based on singular value decomposition~\cite{svd1987}; and in the second method (QT), we use unit quaternions~\cite{horn1987}.} A \Revise{linear} Kalman filter\Revise{\cite{chatelain2013real}} can be used on the needle tip position, for noise reduction, before passing it to either one of the above methods. \Revise{This filter can predict the future state of a system, so it is very useful for the prediction of the position of the needle.}
\section{Experiment}
\label{sec:experiment}
\begin{figure}[h]
\centering
  \includegraphics[width=0.99\columnwidth]{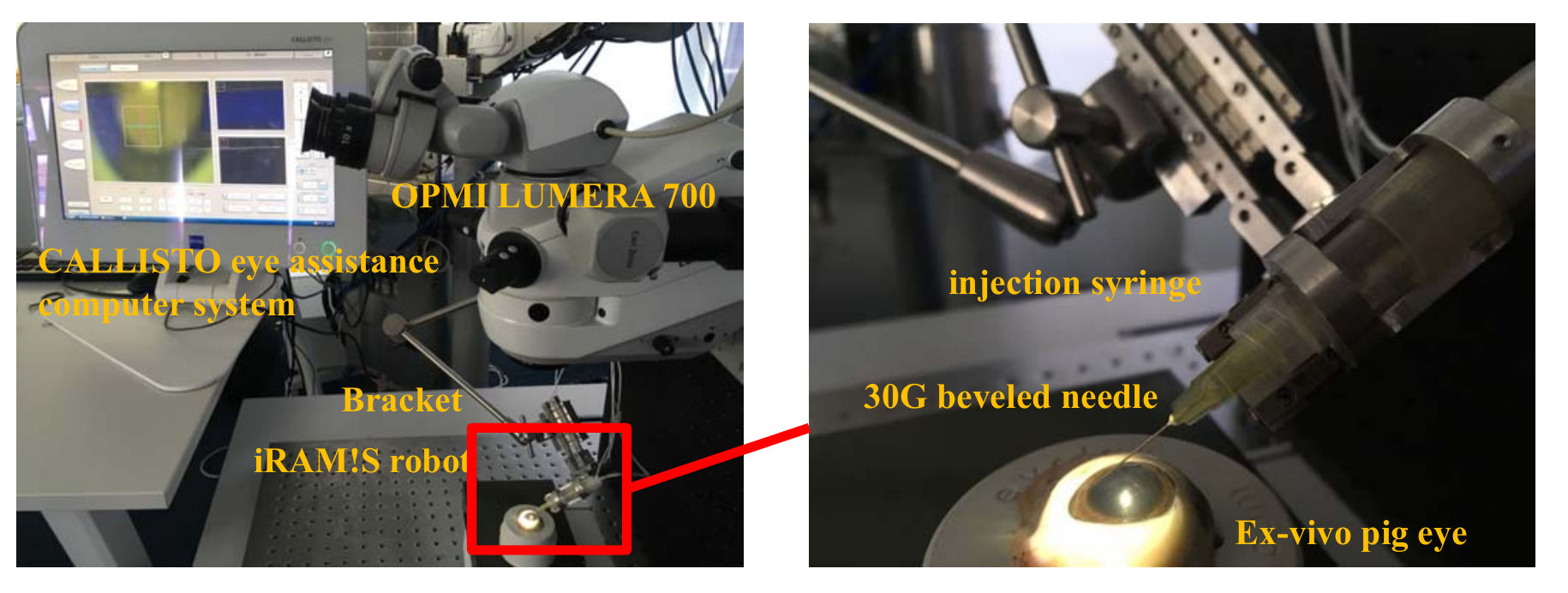}
% figure caption is below the figure
\caption{The experimental setup for hand-eye calibration with ex-vivo pig eye.}
\label{fig:ExprimentSetup}       % Give a unique label
\end{figure}

The experimental setup with an ex-vivo pig eye is depicted in Fig.~\ref{fig:ExprimentSetup}. The OPMI LUMERA 700, with integrated RESCAN 700 OCT engine, is fixed on the optical table to reduce the influence of ambient vibration. The $\mathit{CALLISTO}$ eye assistance computer system (Carl Zeiss Meditec AG., Germany) is utilized to simultaneously display the microscopic (en-face) image and B-scan images. The iRAM!S eye surgical robot with 5 DoF is mounted on the adjusted bracket. Piezo motors (SmarACT GmbH, Germany), used in the robot, provide 1 $\upmu$m of accuracy using PID control with integrated incremental optical encoders. A 30G needle, typical for ophthalmic injection with a diameter of 0.31mm, is mounted on the end effector of the robot. The needle tip is fixed in its initial position very close to the ex-vivo pig eye to simulate an ophthalmic operation. The ex-vivo pig eye is fixed with pins into a rubber support.
The \Revise{program} is executed on the laptop with an Intel Core i7 CPU in Ubuntu 16.04 with Point Cloud Library 1.8 and OpenCV 3.3 in C++.%\cite{rusu20113d}

% \begin{figure*}
% % Use the relevant command to insert your figure file.
% % For example, with the graphicx package use
%   \includegraphics[width=0.9\textwidth]{figures/Ball.pdf}
% % figure caption is below the figure
% \caption{The ball attach on the end of tip in microscopic image (a) and the ball is detected in OCT volume using RANSAC sphere fitting (b).}
% \label{fig:Ball}       % Give a unique label
% \end{figure*}

We are aiming to intraoperatively calibrate the robot and OCT camera with micro-scale movements. In calibration trajectory $\#1$, the needle is moved 10 steps, each being 20 $\upmu$m, first in the X-, then Z-, then Y- direction. Trajectory $\#2$, is a zigzag, in which the needle moves 5 steps, each being  20 $\upmu$m, in the X-, Z- and Y- direction. The same trajectory is repeated once again. After each step is finished, the OCT engine is triggered in order to capture an image volume. The parameters for the OCT calibration are listed in Tab.~\ref{tab:OCTparameters}. These parameter were used to calculate the corrected needle tip position.
%\vspace{-0.1cm}
\begin{table}[htbp]
\centering
\caption{The OCT calibration parameters (mm)}
\label{tab:OCTparameters}       % Give a unique label
\begin{tabular}{llllll}
\hline\noalign{\smallskip}
 $x_c$ & $z_{xc}$ & $y_c$&  $z_{yc}$  \\%& $d_{max}$
\noalign{\smallskip}\hline\noalign{\smallskip}
1.489& 151.563  & 1.068 & 428.541  \\% & 0.013
\noalign{\smallskip}\hline
\end{tabular}
\end{table}

\begin{figure}[htbp]
   \centering
   \subfigure[The ball marker in microscopic view]{
     \includegraphics[width=0.44\columnwidth,height=0.15\textheight]{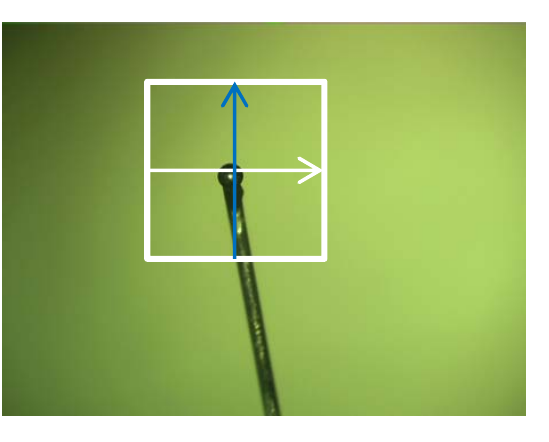}
     \label{fig:Ball_a}
   }
   \subfigure[The ball marker detected in 3D]{
     \includegraphics[width=0.49\columnwidth,height=0.15\textheight]{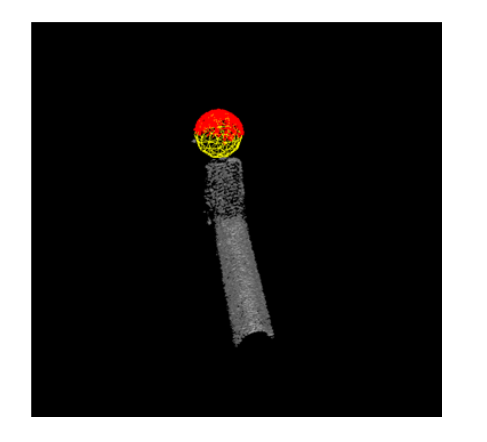}
     \label{fig:Ball_b}
   }
   \caption{The steel ball attached to the tool tip in microscopic image (a) and the detected steel ball in 3D OCT volume using RANSAC sphere fitting (b).}
   \label{fig:Ball}
 \end{figure}

To evaluate the accuracy of our calibration method, we \Revise{compared} it with a reference method (marker-based). The reference method is created by attaching a 0.5mm diameter steel ball (Chrome AISI 52100 Grade 10) with a diameter accuracy of 2.5 $\upmu$m on the end of the needle. The steel ball has a half sphere surface in the OCT volume, which makes it detectable using RANSAC sphere fitting, see Fig.~\ref{fig:Ball}.
  % (%(Fig.~\ref{fig:Ball_a})

\section{Results}
\begin{figure}[htbp]
 \centering
  \subfigure[Trajectory$\#1$ of marker ball center]{
    \includegraphics[width=0.45\columnwidth,height=0.15\textheight]{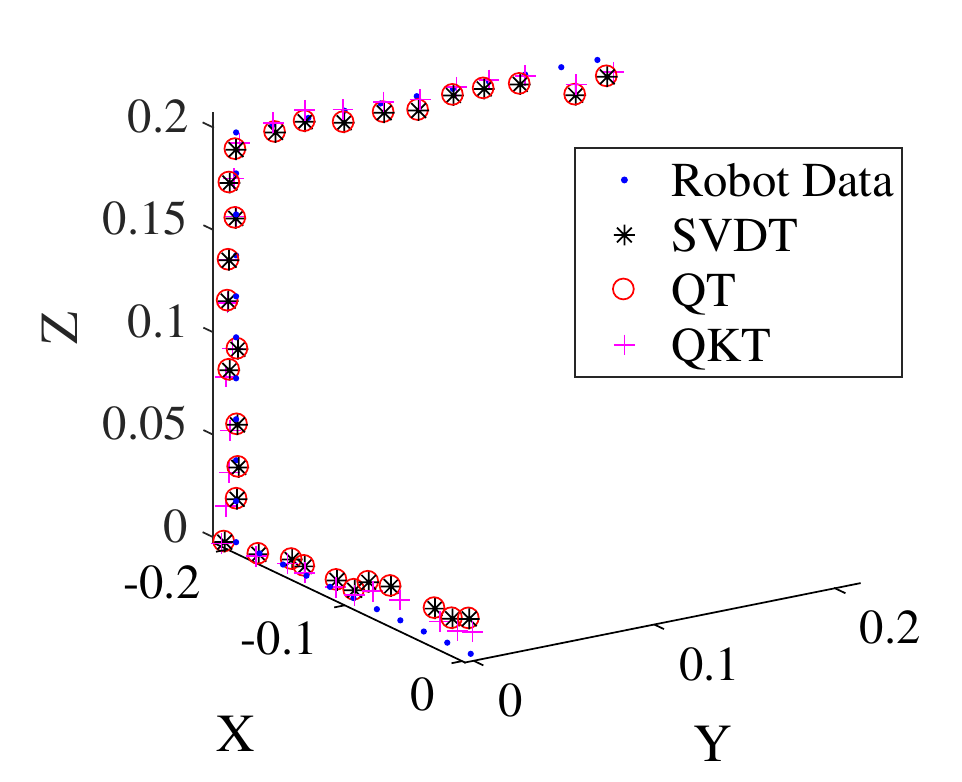}
    \label{fig:dataBall_N}
  }
  \subfigure[Trajectory$\#2$ of marker ball center]{
    \includegraphics[width=0.45\columnwidth,height=0.15\textheight]{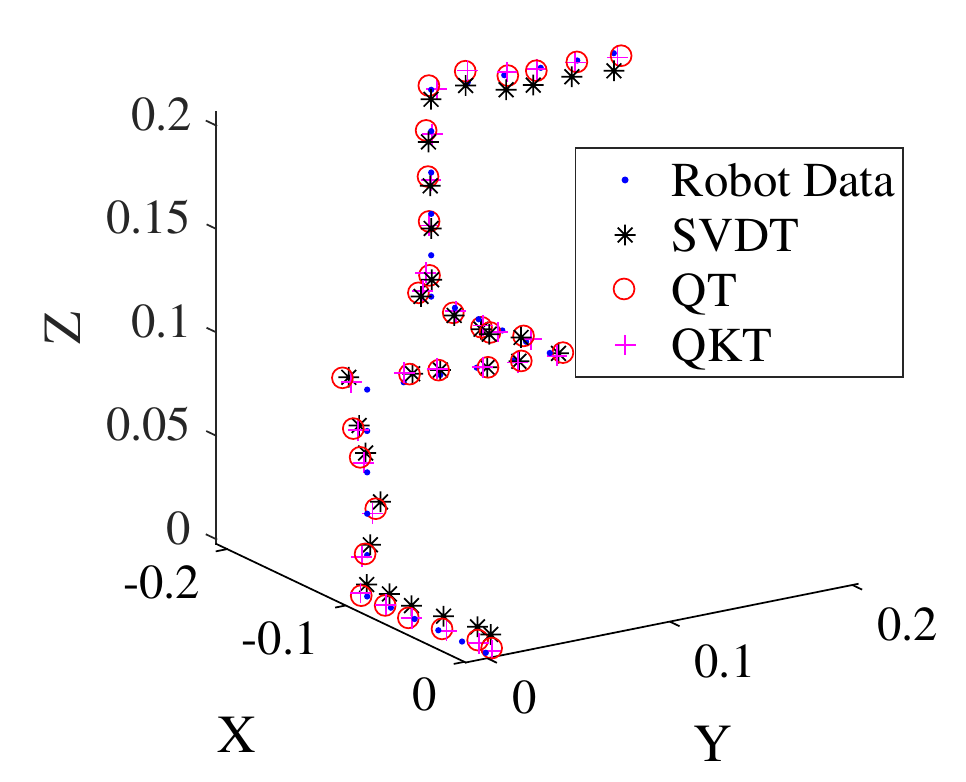}
    \label{fig:dataBall_Z}
  }
  \caption{The marker ball center trajectory$\#1$ (a) and trajectory$\#2$ (b) in robot coordinate system and the data points transformed from OCT volumes using QT, SVDT, and QKT.}
  \label{fig:dataBall}
\end{figure}

\begin{figure}[htbp]
  \centering
  \subfigure[Trajectory$\#1$ of needle tip]{
    \includegraphics[width=0.45\columnwidth,height=0.15\textheight]{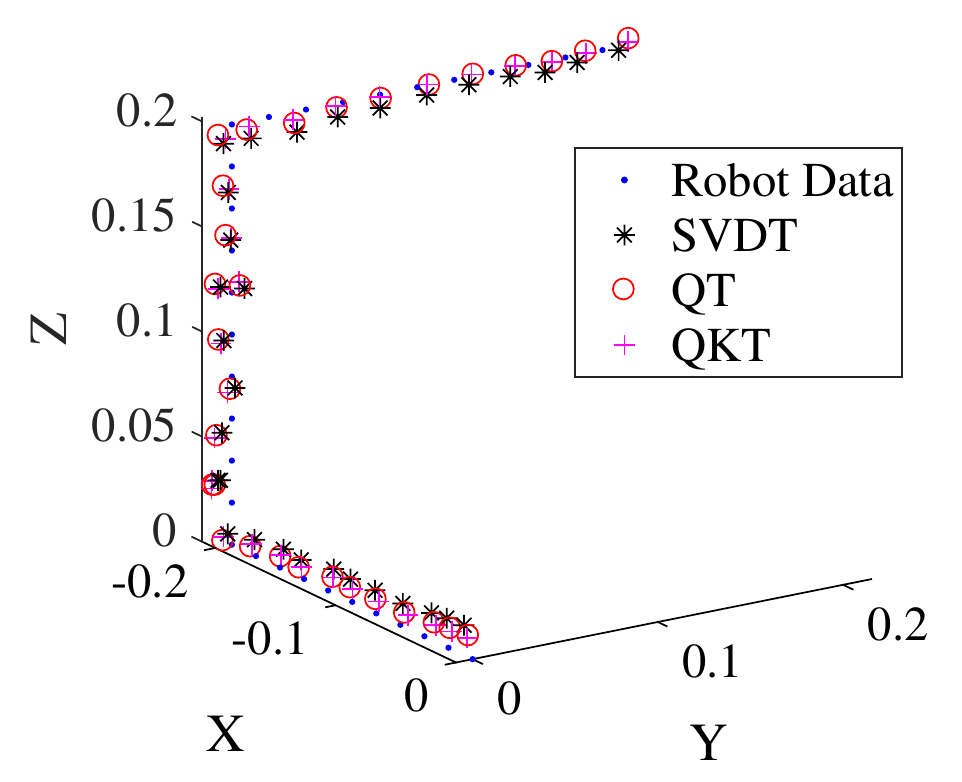}
    \label{fig:dataNeedle_N}
  }
  \subfigure[Trajectory$\#2$ of needle tip]{
    \includegraphics[width=0.45\columnwidth,height=0.15\textheight]{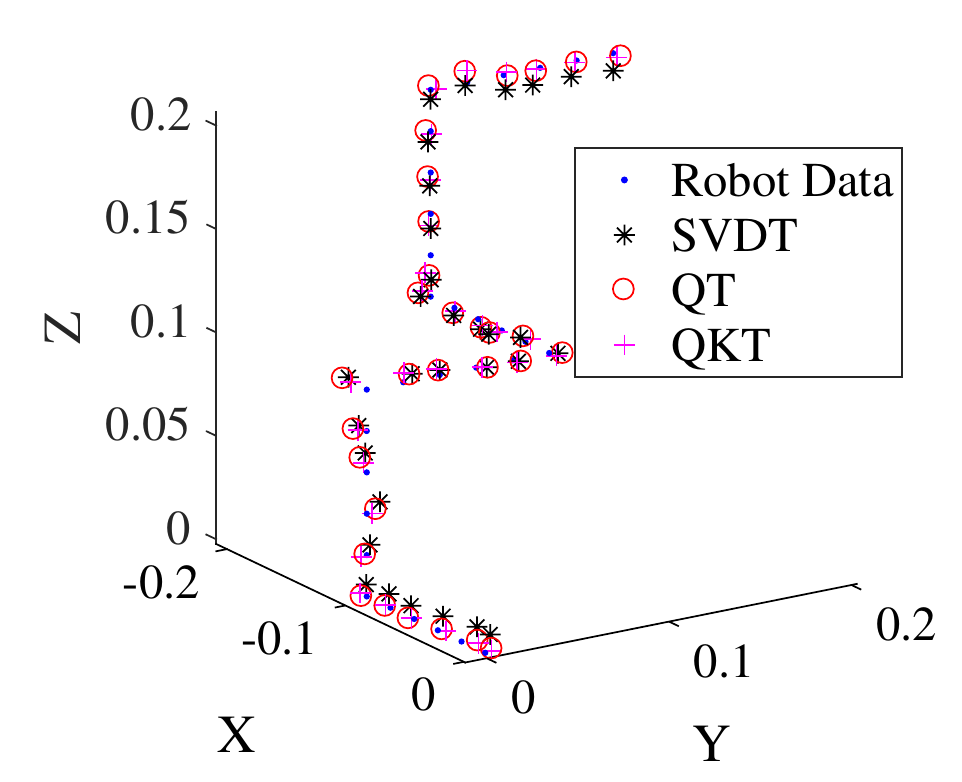}
    \label{fig:dataNeedle_Z}
  }
  \caption{The needle tip trajectory$\#1$ (a) and trajectory$\#2$ (b) in robot coordinate system and the data points transformed from OCT volumes using QT, SVDT, and QKT.}
  \label{fig:dataNeedle}
\end{figure}

%Each trajectory has 31 OCT volumes giving 31 positions of the needle tip or the center of the ball marker. The initial needle tip position is considered as (0,0,0), thus the needle tip psition in robot's coordinate system is obtained from the movement.
Fig.~\ref{fig:dataBall} and Fig.~\ref{fig:dataNeedle} show the trajectories of the marker based method and our proposed calibration method, respectively. The trajectories of both methods are transformed from the OCT coordinate system to the robot coordinate system with a transformation matrix derived from \Revise{the methods described in Section~\ref{sec:methodhand}: SVDT, QT and quaternion method with \Revise{linear} Kalman filter (QKT).} As shown in the figures, all three methods resulted in a trajectory that overall fits well with the trajectory in the robot coordinate system (i.e Robot Data in Fig. 8 and Fig. 9), even though some outliers exist. The average computation time for detecting and transforming one marker ball center and one needle tip position is 1.5s and 0.6s, respectively.
\label{sec:results}
%\vspace{-0.5cm}

The calibration error for a point is calculated as,
\begin{equation}
e_i = \sqrt{(x^R_i-x^{R'}_i)^2+(y^R_i-y^{R'}_i)^2+(z^R_i-z^{R'}_i)^2}
\end{equation} where $(x^R_i,y^R_i,z^R_i)$ is the (needle tip/marker ball center) position in the robot coordinate system and $(x^{R'}_i,y^{R'}_i,z^{R'}_i)$ is the (needle tip/marker ball center)  transformed from the OCT coordinate system.

\begin{figure}[htbp]
  \centering
  \subfigure[Marker ball calibration error]{
    \includegraphics[width=0.47\columnwidth]{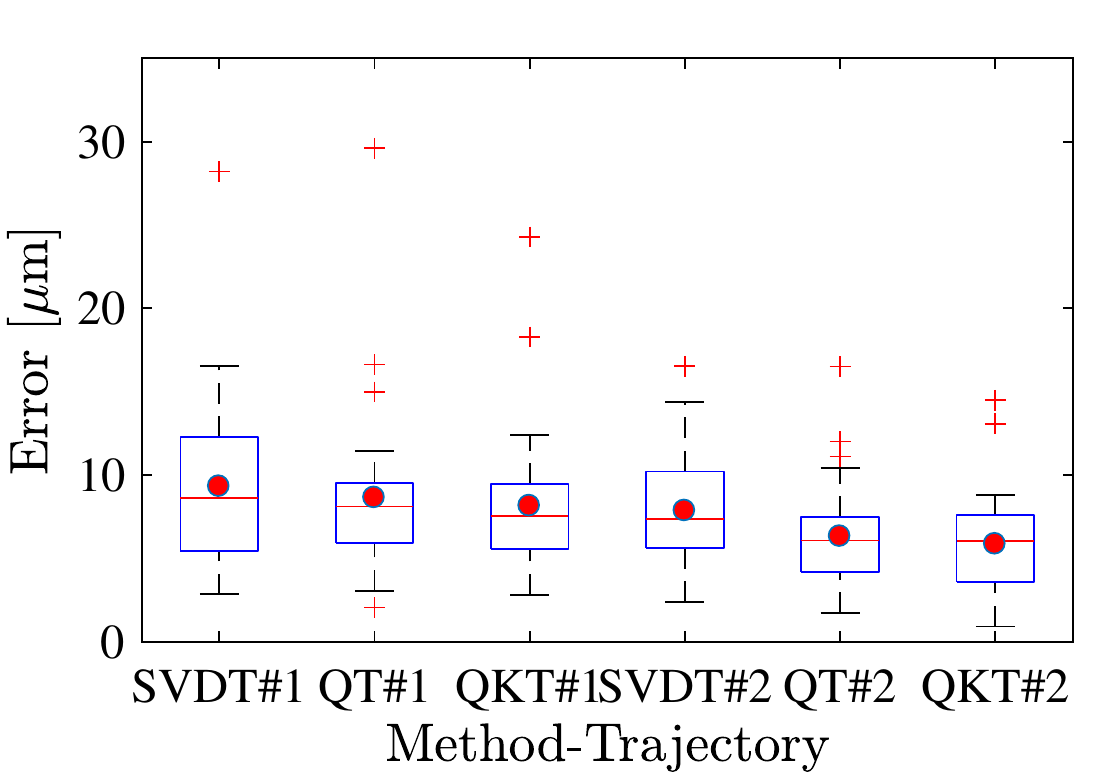}
    \label{fig:dataErrorBall}
  }
  \subfigure[Needle tip calibration error]{
    \includegraphics[width=0.46\columnwidth]{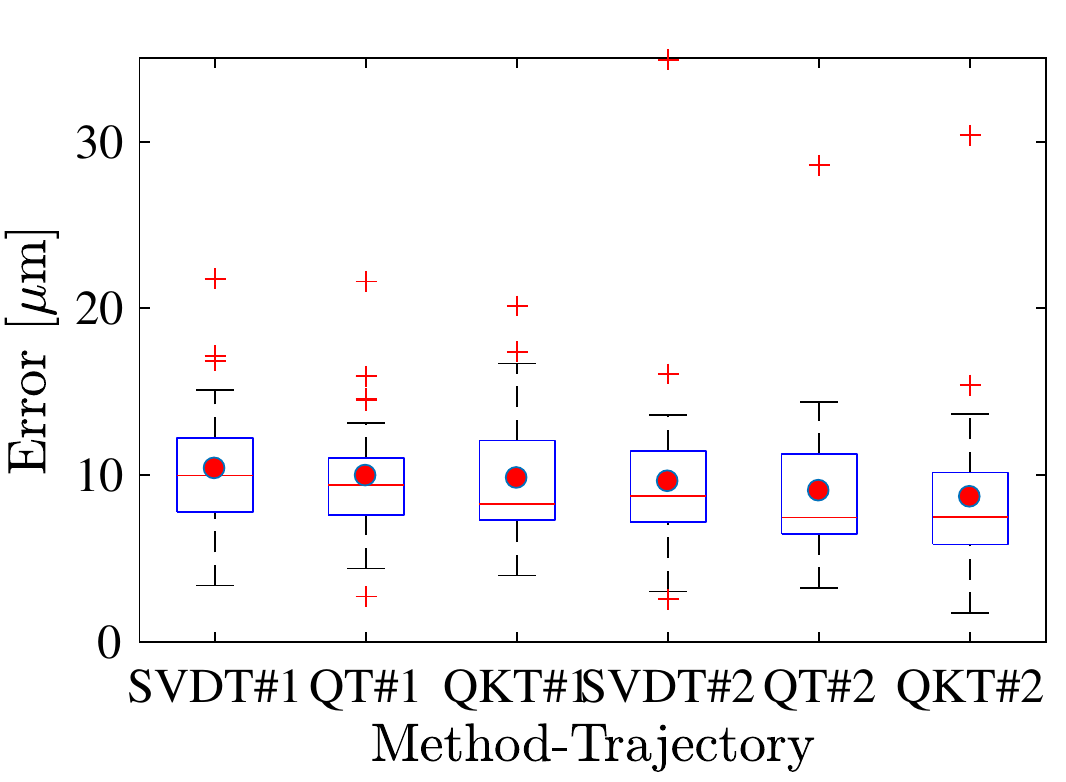}
    \label{fig:dataErrorNeedle}
  }
  \caption{The calibration error for two trajectories with each method using steel ball(a) and needle tip (b).}
  \label{fig:dataError}
\end{figure}
%\vspace{-0.1cm}

Fig.~\ref{fig:dataError} gives the error $e$ with box-and-whisker plots for each method and trajectory. The maximum whisker length is specified as 1.0 times of the interquartile range. Data points beyond the whiskers are treated as outliers. The whiskers are showing the minimum and maximum recorded error while the \Revise{first (25\%) and third (75\%) quartile are shown at the bottom and top edge} of the box. Bands and red dots represent the median and the mean of the error data, respectively. From the figure, we can see that all the methods maintain a low error without a significant difference. For the marker based method, QKT outperforms SVDT $\&$ QT slightly, with an average error of 0.8 $\upmu$m and 1.2 $\upmu$m for trajectory$\#1$ and $\#2$, respectively. For the needle tip method (i.e our proposed method), QKT outperforms the other two methods slightly with an average error of 0.4  $\upmu$m and 0.6 $\upmu$m for trajectory$\#1$ and $\#2$, respectively. It also shows that the performance gap between the marker based method and the needle tip method is very small with an average error of 7.0 $\upmu$m and 9.2 $\upmu$m, respectively.
\begin{figure}[htbp]
  \centering
  \subfigure[Marker based method]{
    \includegraphics[width=0.46\columnwidth]{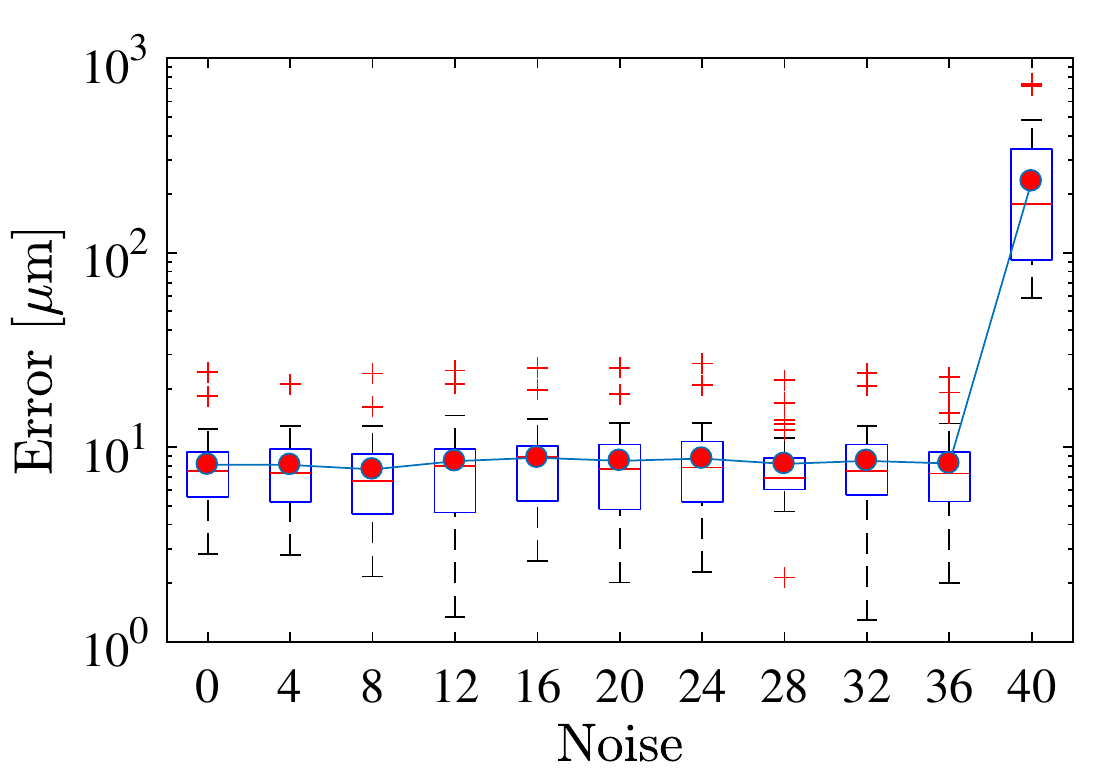}
    \label{fig:dataErrorNeedle}
  }
  \subfigure[Needle tip method]{
    \includegraphics[width=0.46\columnwidth]{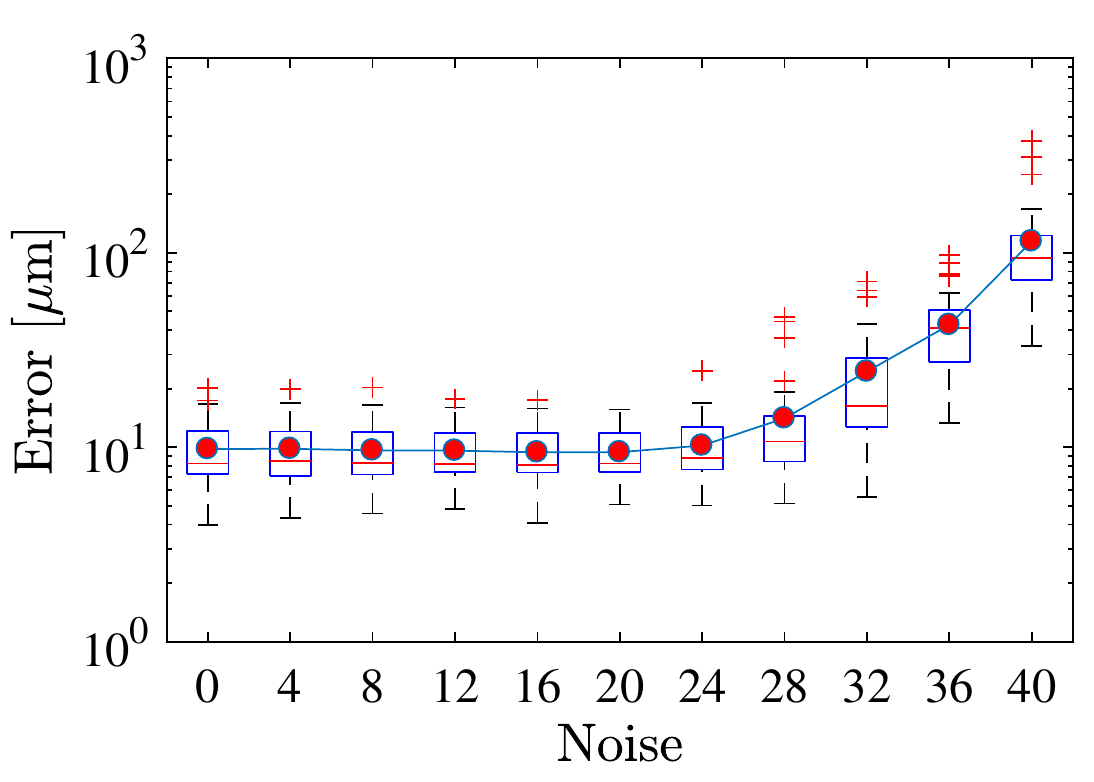}
    \label{fig:dataErrorNeedle2}
  }
  \caption{The relationship between the calibration error for each point and \Revise{standard deviation sigma of the Gaussian noise} for marker based method (a) and needle tip based method (b).}
  \label{fig:Noise}
\end{figure}
%\vspace{-0.1cm}

In order to \Revise{test the robustness} of the proposed method, we carry out a noise test for the QKT method (i.e. method with the best error), trajectory$\#1$. A zero-mean Gaussian noise is added to each B-scan image from standard deviation $\sigma$ = 0 to $\sigma$ = 40 by steps of 4. The relationship between the calibration error and the degree of noise added for both methods is shown in Fig.~\ref{fig:Noise}. We can see that the error for the marker based method (Fig.~\ref{fig:dataErrorNeedle}) remains stable before $\sigma$ = 36, but changes dramatically for $\sigma$ = 40. This is due to the fact, that with too much noise added in the image, the sphere fitting cannot work correctly. For the needle tip method (Fig.~\ref{fig:dataErrorNeedle2}), the error remains stable before $\sigma$ = 24, and steadily increases with more noise. This is because we are detecting shifted versions of the actual needle, due to the noise attached to the tip. From the above results, we find that our proposed method shows a suitable accuracy performance within a clinical tolerance. The method also showed \Revise{robustness} to noise influence.

  %%It is said that the ideal needle position accuracy for ophthalmic surgery is 10 $\upmu$m~\cite{gijbels2014design}.

\section{Conclusion\&Discussion}
\label{sec:conculsion}
We proposed a flexible framework for calibrating an ophthalmic robot with an intra-operative MI-OCT. No markers were introduced in the calibration process, which makes our method easier to integrate into the current operation room, by avoiding instrument modification and sterilization. The calibration error of 9.2 $\upmu$m meets the requirement of ophthalmic operations. \Revise{Although micron-level displacements in the calibration process may cause safety concerns}, the relative position between needle tip and eye tissue surface can be estimated to ensure safety. Accordingly, systems-theoretic safety assessments\Revise{\cite{systemsafty}} could be a potential path for completing, refining, adapting, and customizing the whole setup and framework.

%Currently, the calibration error of proposed method is depended on the needle tip accuracy which is ensured by the high resolution of OCT volumetric images. The reason is that the needle tip is one of the biggest interest in ophthalmic surgery which related to safety concerns.%

\Revise{In future work, besides the needle tip information, the extra geometric constraints from the needle\Revise{~\cite{vasconcelos2016spatial}} or even the full geometric information of needle (i.e. CAD model of the needle) could be a potential improvement to increase the accuracy and robustness of the method. Nevertheless, this paper firstly verify the feasibility and accuracy performance of hand-eye calibration between OCT camera and ophthalmic robot, which motivates future work of path planning for ophthalmic injection and OCT based visual servoing control to achieve an enhanced RAS.}

%%%%%%%%%%%%%%%%%%%%%%%%%%%%%%%%%%%%%%%%%%%%%%%%%%%%%%%%%%%%%%%%%%%%%%%%%%%%%%%%
%\section*{APPENDIX}
%
%Appendixes should appear before the acknowledgment.
%
%\section*{ACKNOWLEDGMENT}
%
%The preferred spelling of the word ÒacknowledgmentÓ in America is without an ÒeÓ after the ÒgÓ. Avoid the stilted expression, ÒOne of us (R. B. G.) thanks . . .Ó  Instead, try ÒR. B. G. thanksÓ. Put sponsor acknowledgments in the unnumbered footnote on the first page.

%%%%%%%%%%%%%%%%%%%%%%%%%%%%%%%%%%%%%%%%%%%%%%%%%%%%%%%%%%%%%%%%%%%%%%%%%%%%%%%%

\bibliographystyle{IEEEtran}
{\scriptsize
\vspace{0.01 cm}
\bibliography{MyCollection}
}

\end{document}